\pdfoutput=1
\documentclass{article}

\usepackage[preprint]{corl_2026} % arXiv version: non-anonymous

\title{HALO-WA: Hybrid-Attention Latent-Guided Online Reinforcement Learning for World-Action Models}

\author{
\textbf{Angen Ye}$^{1,2,*}$ \quad
\textbf{Weijie Ke}$^{3,*}$ \quad
\textbf{Xiaofeng Wang}$^{3,4}$ \quad
\textbf{Xinze Chen}$^{3}$ \quad
\textbf{Chaojun Ni}$^{3}$ \\
\textbf{Guosheng Zhao}$^{3}$ \quad
\textbf{Boyuan Wang}$^{1,2}$ \quad
\textbf{Zheng Zhu}$^{3,\dagger}$ \quad
\textbf{Junjie Xie}$^{1,2}$ \quad
\textbf{Dapeng Zhang}$^{1,2}$ \\[1mm]
$^{1}$Institute of Automation, Chinese Academy of Sciences, Beijing, China \\
$^{2}$School of Artificial Intelligence, University of Chinese Academy of Sciences, Beijing, China \\
$^{3}$GigaAI, Beijing, China \\
$^{4}$Tsinghua University, Beijing, China \\
$^{*}$Equal contribution \\
$^{\dagger}$Corresponding author: \texttt{zhengzhu@ieee.org}
}

\usepackage{amsmath}
\usepackage{amssymb}
\usepackage{algorithm}
\usepackage{algpseudocode}

\usepackage{booktabs}
\usepackage{multirow}
\usepackage{makecell}
\usepackage{graphicx}
\usepackage{wrapfig}
\usepackage{caption} 
\begin{document}
\maketitle

%===============================================================================
\vspace{-2em}

\begin{abstract}
World-action (WA) models can generate long-horizon action chunks for general-purpose robotic manipulation, but they remain vulnerable to calibration, perception, and contact-dynamics errors in real-world precision tasks, often failing in the final few millimeters of alignment or insertion. We propose HALO-WA, a hybrid-attention latent-guided online reinforcement learning (RL) framework for WA models, which leverages latent features and action priors from the WA generation process through a lightweight actor-critic adapter to enable fast online adaptation to real deployment errors. HALO-WA introduces a hybrid-attention structure that preserves the temporal consistency of action chunks while reading task-relevant information from WA latents conditioned on visual context and end-stage correction requirements, thereby producing refined action chunks. We validate HALO-WA on four real-world precision manipulation tasks, where it improves the average success rate from 26.4\% for WA-base to 87.1\%, outperforming the strongest baseline by 19.2 percentage points while requiring only 45--75 minutes of online training per task. To facilitate reproducibility, we further conduct supplementary simulation experiments in RoboTwin and release the code at \url{https://github.com/YeanRoot/HALO-WA}.
\end{abstract}
% \vspace{-1em}

%===============================================================================
% \vspace{-1.5em}

\section{Introduction}
\vspace{-0.5em}
World-action (WA) models provide a promising route toward general-purpose robotic manipulation. They can generate long-horizon action chunks conditioned on visual observations, robot states, and task specifications\cite{wang2026world}\cite{hou2026world}\cite{team2025gigaworld}\cite{kim2026cosmos}\cite{bi2025motus}. However, deploying such models on real robots remains challenging. In precision manipulation, small errors in calibration, perception, control, or contact dynamics can be amplified during the final execution stage. As a result, even if a WA model generates a seemingly reasonable action sequence, it may still fail in the last few millimeters of insertion, alignment, or contact. This precision deployment gap is difficult to eliminate through offline training alone, because the most critical errors are often tied to a specific robot, a specific environment, and real physical interaction.

Online reinforcement learning offers a potential solution to this problem\cite{luo2025precise}\cite{lei2025rl}\cite{yuan2024learning}. By practicing in the target environment, a robot can learn from real successes, failures, and human intervention feedback, with a particular focus on improving the contact-rich stages where offline policies are most fragile. However, directly applying online reinforcement learning to a large-scale WA model is impractical: full-parameter updates are computationally expensive, slow to adapt, and prone to overfitting under limited real-robot interaction data. Training a small policy from scratch is also undesirable, because it discards the visual, temporal, and behavioral priors already encoded in the WA model. Therefore, the key question is: how can we enable fast online reinforcement learning for precision control while preserving the general capability of a pretrained WA model?

Recent studies have shown that frozen robotic foundation models can be combined with lightweight online reinforcement learning to improve real-world deployment performance. GR-RL\cite{li2025gr} shows that online RL can specialize a general vision-language-action (VLA) policy into a specialist policy for long-horizon precision manipulation. RL Token\cite{xu2026rl} further demonstrates that compact representations extracted from a frozen VLA model, together with VLA reference actions, can drive a small actor-critic policy. However, these methods are mainly designed for VLA-style policy interfaces\cite{ye2025vla}\cite{lu2025vla}\cite{black2024pi_0}\cite{team2025gigabrain}, such as policy fine-tuning, action residual learning, or compact token readout. WA models have a different action generation structure: task-relevant information required for end-stage alignment, contact, and insertion is often distributed in variational autoencoder (VAE) or world-action latent features\cite{ye2026world}\cite{yuan2026fast}\cite{feng2026harmowam}\cite{liu2026oa}, rather than being fully captured by the final action output. Therefore, online RL for WA models requires a WA-native adaptation mechanism that preserves the WA reference action prior while exploiting distributed WA latent features for precise action correction.
\vspace{-0.5em}
\begin{center}
    \includegraphics[width=0.95\textwidth]{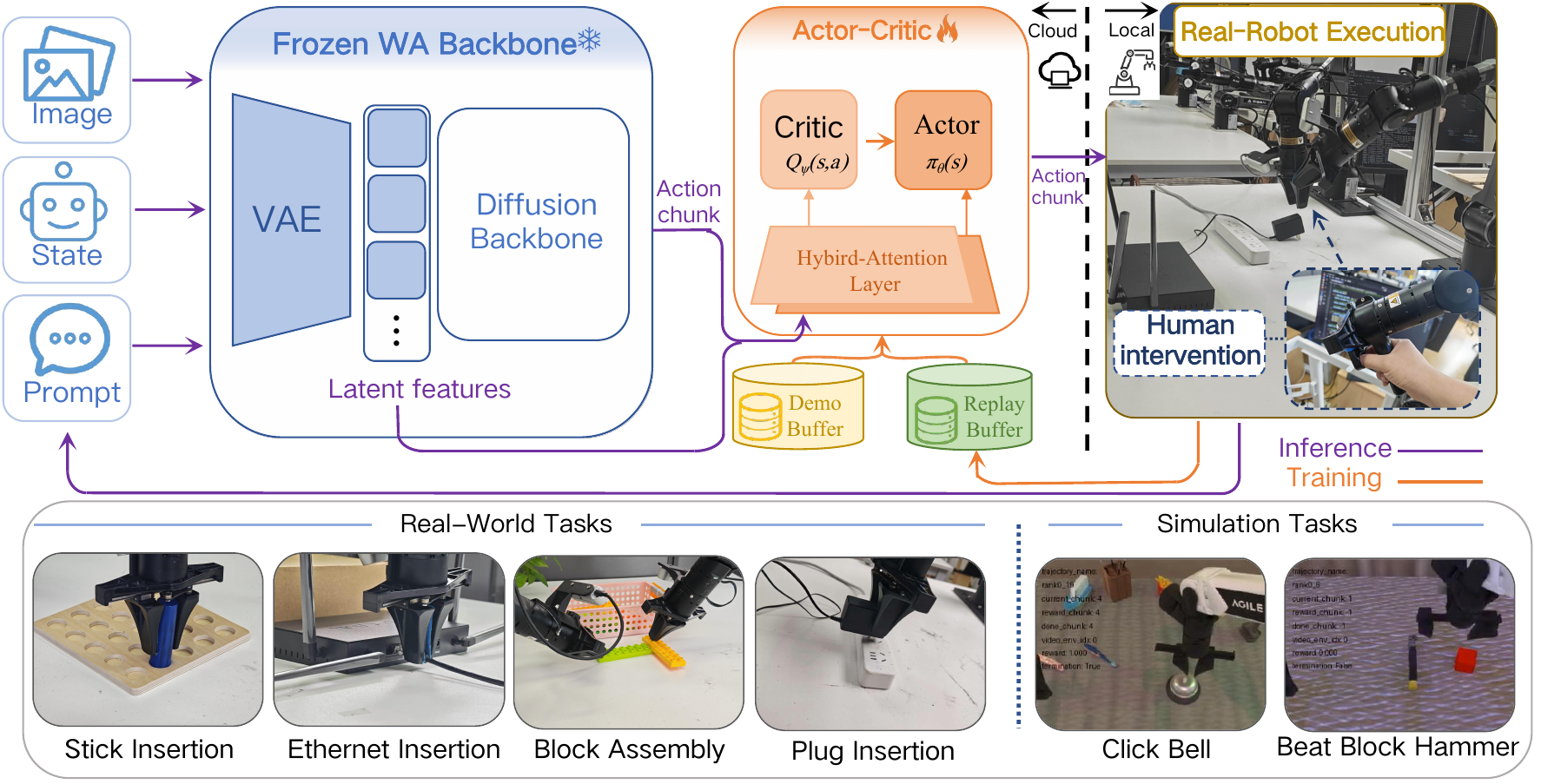}
    \captionof{figure}{
    Overview of HALO-WA. The figure illustrates the cloud-robot online adaptation loop, including WA-guided action refinement, real-robot execution with human intervention, and evaluation on real-world and RoboTwin precision manipulation tasks.
    }
    \label{fig:overview}
\end{center}
\vspace{-0.5em}
We propose HALO-WA: Hybrid-Attention Latent-Guided Online Reinforcement Learning for World-Action Models. HALO-WA freezes the pretrained WA backbone and trains a lightweight actor-critic adapter for online action refinement. The adapter is mainly guided by the WA reference action chunk and latent features extracted from the WA generation process, with the current robot state used as supplementary context. The reference action provides a strong action prior, while the latent features expose the internal world-action representations of the WA model\cite{jha2026reconstruction}. A hybrid-attention module fuses these inputs and predicts a refined action chunk\cite{vaswani2017attention}\cite{bahdanau2014neural}, enabling the policy to follow the WA prior when it is reliable and deviate from it when precise correction is needed.

As shown in Fig.~\ref{fig:overview}, HALO-WA adopts a cloud-robot collaborative closed-loop training architecture. The cloud side performs WA inference, adapter forward inference, and online updates, and generates refined action chunks based on the observations and states returned from the robot. The robot side only executes the actions sent from the cloud, detects human intervention, and sends back new interaction data. During online execution, all actually executed actions are stored in the replay buffer, including both policy actions and human intervention actions, providing real interaction samples for subsequent actor-critic updates. Through this closed loop, HALO-WA can continuously collect feedback during real-world deployment and rapidly adapt to end-stage alignment, contact, and insertion errors.

We evaluate HALO-WA on challenging precision manipulation tasks. In the real world, we construct four tasks that require accurate end-stage contact and alignment: stick insertion, Ethernet insertion, block assembly, and power plug insertion. To facilitate reproducibility, we further implement two tasks in the RoboTwin\cite{chen2025robotwin} simulation environment: Click Bell and Beat Block Hammer.

% \vspace{+1em}

The main contributions of this paper are summarized as follows:
\begin{itemize}
    \item We propose \textbf{HALO-WA, a latent-guided online RL framework} that refines frozen WA models with a lightweight actor-critic adapter for real-world precision manipulation.
    \item We design a \textbf{hybrid-attention actor-critic adapter} that combines WA reference actions and distributed WA latent features for stable chunk-level action correction.
    \item We validate HALO-WA on four real-world tasks and two RoboTwin tasks, \textbf{improving the average real-world success rate from 26.4\% to 87.1\% with only 45--75 minutes of online training per task}.
\end{itemize}
% The main contributions of this paper are threefold: 
% (1) We propose HALO-WA, a latent-guided online RL framework that refines frozen WA models with a lightweight actor-critic adapter for real-world precision manipulation. 
% (2) We design a hybrid-attention actor-critic adapter that combines WA reference actions and distributed WA latent features for stable chunk-level action correction. 
% (3) We validate HALO-WA on four real-world tasks and two RoboTwin tasks, improving the average real-world success rate from 26.4\% to 87.1\% with only 45--75 minutes of online training per task.
\vspace{-1em}

%===============================================================================
\section{Related Work}
% \vspace{-1em}
% World-action (WA) models have evolved from utilizing generative video priors for robotic control to architectures that jointly model future world states and action generation. Recent research further explores efficient inference, causal structure, and long-horizon planning, highlighting the growing role of WA models as general visuomotor policy backbones. However, most existing WA models are exclusively trained offline and cannot correct deployment-specific errors caused by calibration, contact dynamics, or terminal alignment. HALO-WA addresses this precision deployment gap by introducing online reinforcement learning to refine a frozen WA policy using real-interaction feedback without updating the large backbone.
\vspace{-0.5em}
\subsection{World-Action Models}
\vspace{-1em}
World-action (WA) models couple video world modeling with action generation to produce long-horizon action chunks from visual observations, robot states, and task specifications.
Cosmos Policy~\cite{kim2026cosmos} fine-tunes large-scale video diffusion models for visuomotor control and planning, demonstrating that generative video priors can transfer to robotic action.
GigaWorld-Policy~\cite{ye2026gigaworld} proposes an efficient action-centered architecture that integrates world state prediction with action generation at scale.
Fast-WAM~\cite{yuan2026fast} revisits the role of test-time future imagination in WA inference and shows that a streamlined scheme can match the performance of full imaginative rollouts.
Causal world modeling~\cite{li2026causal} incorporates causal structure into the world model to improve physical generalization across novel interaction scenarios.
Despite their expressive visual priors and long-horizon planning capability, these models are trained exclusively offline and cannot adapt to deployment-specific errors.
Inaccuracies in robot calibration, contact dynamics, or terminal alignment accumulate during the execution stage and are not corrected after training\cite{wang2025embodiedreamer}, creating a precision deployment gap.
HALO-WA targets this gap by introducing online reinforcement learning that refines a frozen WA policy using real-interaction feedback without updating the large backbone.
\vspace{-0.5em}
\subsection{Online Reinforcement Learning and Interactive Imitation Learning}
\vspace{-1em}
Online learning and interactive imitation learning methods improve robot policies through environment interaction and expert feedback.
DAgger~\cite{pmlr-v15-ross11a} establishes the theoretical foundation for interactive imitation learning: by iteratively running the current policy and querying an expert for corrective labels, it provably reduces the distribution shift inherent to offline behavioral cloning.
HG-DAgger~\cite{kelly2019hg} extends this framework with human-gated intervention, allowing an expert to selectively override the policy in unsafe or suboptimal states.
HIL-SERL~\cite{luo2025precise} combines human intervention with reward-based online reinforcement learning, achieving precise and dexterous manipulation through an actor-critic loop with real-time human corrections.
RL Token~\cite{xu2026rl} bootstraps online reinforcement learning on frozen vision-language-action models by compressing the internal representations into a compact token, which drives a small actor-critic policy alongside the reference action of the vision-language-action model.
Together, these methods demonstrate that online interaction effectively closes the gap between offline training and deployment precision.\cite{ye2026fully}
However, they present two complementary limitations for the adaptation of WA models: methods such as HIL-SERL and DAgger ignore pretrained WA priors, discarding the visual, temporal, and behavioral knowledge encoded in the WA backbone; RL Token adopts a compact-token interface designed for transformer-based vision-language-action models, potentially losing the fine-grained spatial and action-generation cues distributed across the latent features of WA models.
HALO-WA addresses both limitations by preserving the priors of the WA model through reference action guidance and distributed latent feature extraction, while enabling stable online improvement through hybrid attention fusion and behavior-cloning regularization.

\section{Method}
\vspace{-1em}

We propose HALO-WA, which learns a lightweight online actor-critic adapter on top of a pretrained world-action model for real-world precision manipulation. The frozen WA backbone provides a reference action chunk and WA visual latent features, and the adapter uses these priors, conditioned on the current robot state, to generate a refined action chunk. During online execution, all actually executed actions, including human interventions, are stored in the replay buffer for actor-critic updates. The system follows a cloud-robot collaborative setup, where the cloud handles WA inference and online learning, while the robot executes actions and returns observations.

\begin{figure*}[t]
    % \vspace{-1em}
    \centering
    \includegraphics[width=0.95\textwidth]{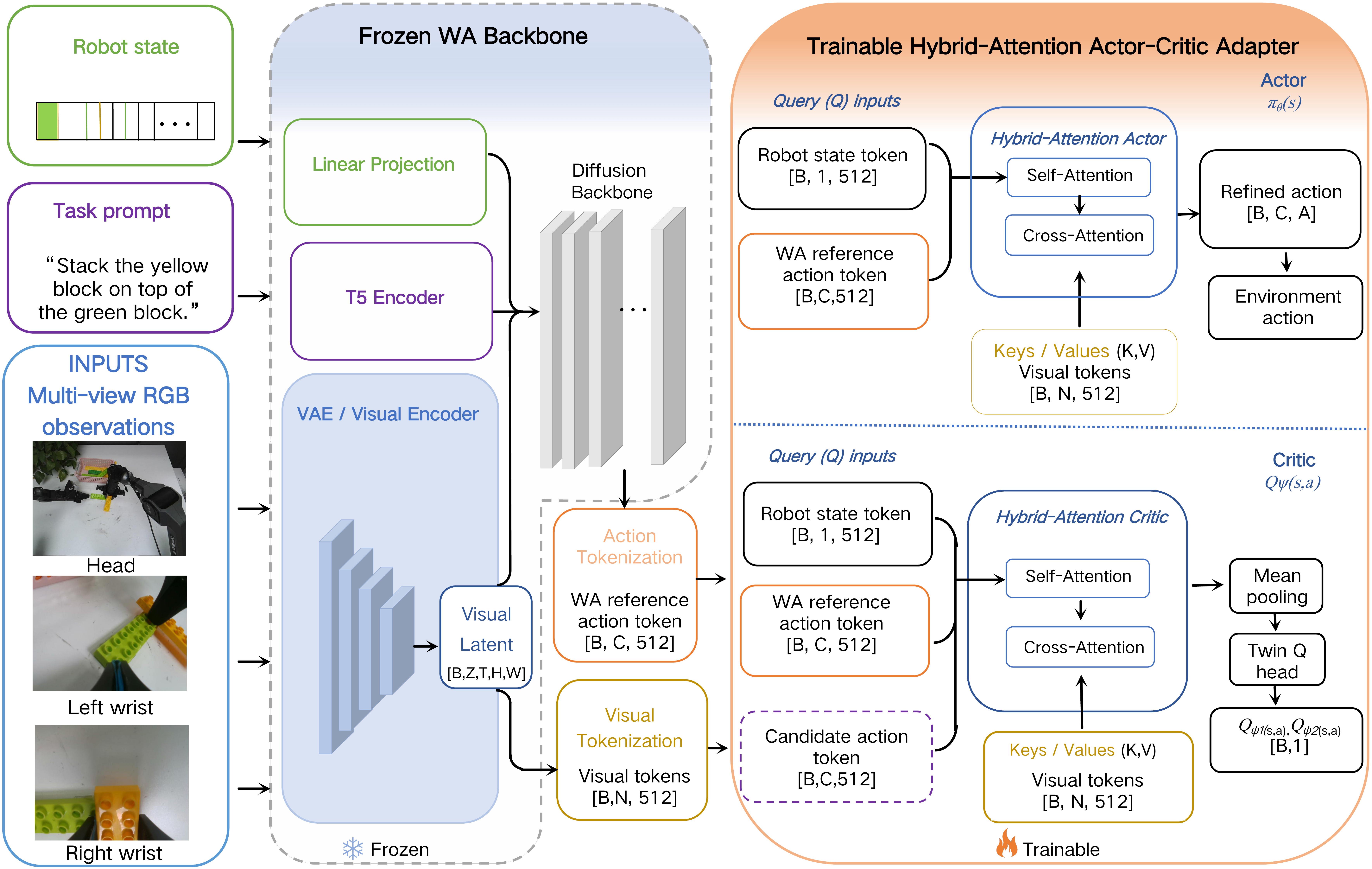}
    \caption{
Hybrid-attention actor-critic adapter in HALO-WA. The figure shows how WA reference actions, robot states, and visual latent features are organized into query-side tokens and latent memory for actor and critic learning.
}
    \label{fig:method}
    % \vspace{-1em}
\end{figure*}

\vspace{-0.5em}
\subsection{Overall Framework}
\vspace{-1em}
Let the frozen WA model be denoted as $\pi_{\mathrm{WA}}$. At time step $t$, the robot receives multi-view observations $o_t$, the narrow robot state $s_t^r$, and the task instruction $p$. Here, $s_t^r$ denotes the proprioceptive robot state, such as joint states, end-effector states, or gripper states. The WA model first generates a reference action chunk of length $H$:
\vspace{-0.5em}
\begin{equation}
\tilde{\mathbf{a}}_{t:t+H-1}
=
\pi_{\mathrm{WA}}(o_t, s_t^r, p).
\end{equation}
% \vspace{-0.3em}
Here, $\tilde{\mathbf{a}}_{t:t+H-1}$ denotes the original action prior produced by the WA model. Meanwhile, we extract latent features from the visual encoding and action generation process of the WA model, denoted as $\mathbf{Z}_t^{\mathrm{WA}}$.

HALO-WA does not directly modify the WA model. Instead, it learns a lightweight adapter on top of the WA output. Specifically, the adapter operates on the first $C$ actions of the reference action chunk, where $C \leq H$. To distinguish the narrow robot state from the reinforcement learning state, we denote the full state observed by the actor-critic adapter as $s_t$:
\vspace{-0.5em}
\begin{equation}
s_t
=
\left(
s_t^r,\,
\tilde{\mathbf{a}}_{t:t+C-1},\,
\mathbf{Z}_t^{\mathrm{WA}}
\right).
\end{equation}
% \vspace{-1em}
That is, $s_t$ is not only the proprioceptive robot state, but a state representation composed of the robot state, the WA reference action, and the WA latent features. The actor outputs a refined action chunk conditioned on this state:
\vspace{-0.5em}
\begin{equation}
\mathbf{a}^{\pi}_{t:t+C-1}
=
\pi_{\theta}(s_t),
\qquad
\mathbf{a}^{\pi}_{t:t+C-1}
\in
\mathbb{R}^{C \times A},
\end{equation}
% \vspace{-1em}
where $A$ is the single-step action dimension. This design enables HALO-WA to preserve the global action prior of the WA model while using the fine-grained spatial and action-generation information contained in WA latent features to refine end-stage actions.
\vspace{-0.5em}
\subsection{Hybrid-Attention Actor-Critic Adapter}
\vspace{-1em}

To enable latent-guided action refinement, HALO-WA uses a lightweight hybrid-attention actor-critic adapter to refine the WA output online. As shown in Fig.~\ref{fig:method}, the adapter encodes the robot state and WA reference action as query-side state-action tokens, denoted as $q_t=[e_t^r,E_{a,t}^{\mathrm{WA}}]$, and encodes the WA visual latent features as latent memory tokens, denoted as $m_t=f_{\mathrm{vis}}(Z_t^{\mathrm{WA}})$. Here, $q_t$ describes the current execution state and action prior, while $m_t$ provides visual and action-generation information from the WA generation process.

For the actor, the state-action tokens first pass through self-attention\cite{vaswani2017attention} to model the structural relationship between the robot state and the WA reference action chunk, preserving the temporal consistency of the refined action chunk. The actor then uses cross-attention\cite{bahdanau2014neural} with $q_t$ as queries and $m_t$ as keys/values to selectively read latent information relevant to the current action correction, such as target position, object geometry, and potential contact regions. The actor finally outputs the refined action chunk $\mathbf{a}^{\pi}_{t:t+C-1}=\pi_{\theta}(s_t)$.

The critic uses the same state representation but additionally receives a candidate action chunk $\mathbf{a}_{t:t+C-1}$ and estimates its value as $Q_{\psi}(s_t,\mathbf{a}_{t:t+C-1})$. Structurally, the critic takes the candidate action together with the state-action tokens as query-side inputs, uses self-attention to model the relationship among the state, reference action, and candidate action, and then reads the WA latent memory through cross-attention. The resulting representation is passed to twin Q heads for value estimation, which mitigates Q-value overestimation in TD3\cite{fujimoto2018addressing}.

Through this query-memory hybrid-attention design, the actor and critic can dynamically retrieve task-relevant information from WA latents while preserving the action prior provided by the WA reference action. When the WA output is reliable, the adapter tends to stay close to the original action; when end-stage alignment, contact, or insertion errors occur, it generates targeted refined action chunks. This preserves the long-horizon action capability of the WA model while enhancing local correction for real-world precision deployment.
\vspace{-0.5em}
\subsection{Online Training Objective}
\vspace{-1em}
HALO-WA trains the actor-critic adapter using TD3 with behavior regularization. Each transition contains the current state $s_t$, the actually executed action chunk $\mathbf{a}_{t:t+C-1}$, reward, next state $s_{t+C}$, and terminal flag. If human intervention occurs during execution, the human action overrides the actor output and is stored as the actually executed action in the replay buffer, ensuring consistency between the stored action and the real state transition.

The critic is trained with a chunk-level TD target. Given a transition sampled from the replay buffer, the target actor first generates the next action chunk under the next state:
\vspace{-0.5em}
\begin{equation}
\mathbf{a}'_{t+C:t+2C-1}
=
\pi_{\theta'}(s_{t+C}).
\end{equation}
% \vspace{-1em}
The TD target is then constructed as:
\vspace{-0.5em}
\begin{equation}
y_t
=
R_t
+
\gamma^C
\min_{j=1,2}
Q_{\psi'_j}
\left(
s_{t+C},
\mathbf{a}'_{t+C:t+2C-1}
\right),
\end{equation}
% \vspace{-1em}
where $R_t$ is the accumulated reward within the current action chunk, $\gamma$ is the discount factor, and $\psi'_j$ denotes the target critic. The critic loss is:
\vspace{-0.5em}
\begin{equation}
\mathcal{L}_{Q}
=
\mathbb{E}_{\mathcal{B}}
\left[
\sum_{j=1}^{2}
\left(
Q_{\psi_j}
\left(
s_t,
\mathbf{a}_{t:t+C-1}
\right)
-
y_t
\right)^2
\right].
\end{equation}
% \vspace{-1em}
The actor objective consists of two terms: one maximizes the action value estimated by the critic, and the other uses behavior cloning (BC) regularization to prevent the actor from drifting away from reliable action distributions in the low-data online learning stage:
\vspace{-0.2em}
\begin{equation}
\mathcal{L}_{\pi}
=
-
\mathbb{E}_{\mathcal{B}}
\left[
Q_{\psi_1}
\left(
s_t,
\pi_{\theta}(s_t)
\right)
\right]
+
\lambda_{\mathrm{BC}}
\mathbb{E}_{\mathcal{B}}
\left[
\left\|
\pi_{\theta}(s_t)
-
\mathbf{a}^{\mathrm{ref}}_{t:t+C-1}
\right\|_2^2
\right].
\end{equation}
% \vspace{-1em}
Here, $\mathbf{a}^{\mathrm{ref}}_{t:t+C-1}$ can come from the WA reference action, the actually executed action in successful trajectories, or human intervention actions. The TD3 term enables the actor to improve from real success and failure feedback, while the behavior cloning regularization prevents the actor from drifting into unsafe or unreasonable action regions when the critic is still inaccurate. In this way, HALO-WA preserves the WA action prior while gradually learning precise corrections that can outperform the original WA output.

% \begin{algorithm}[t]
% \caption{HALO-WA Online Training}
% \label{alg:halo-wa}
% \begin{algorithmic}[1]
% \Require Frozen WA model $\pi_{\mathrm{WA}}$, actor $\pi_{\theta}$, critics $Q_{\psi_1},Q_{\psi_2}$, replay buffer $\mathcal{B}$
% \State Optionally warm up the actor with offline supervised fine-tuning
% \For{each online episode}
%     \State Observe multi-view images $o_t$, robot state $s_t$, and task prompt $p$
%     \State Query the frozen WA model to obtain $\tilde{\mathbf{a}}_{t:t+H-1}$ and $\mathbf{Z}^{\mathrm{WA}}_t$
%     \State Tokenize $\mathbf{Z}^{\mathrm{WA}}_t$ into WA latent tokens
%     \State Predict refined action chunk $\mathbf{a}_{t:t+C-1}=\pi_{\theta}(\mathbf{Z}^{\mathrm{WA}}_t,\tilde{\mathbf{a}}_{t:t+C-1},s_t)$
%     \State Execute the action chunk unless overwritten by human intervention
%     \If{human intervention occurs}
%         \State Store human commands as the executed actions
%         \State Re-query the WA model after intervention to obtain a new reference chunk
%     \EndIf
%     \State Obtain sparse success/failure reward from the human supervisor
%     \State Store transition in replay buffer $\mathcal{B}$
%     \For{each gradient update}
%         \State Sample a batch from $\mathcal{B}$
%         \State Update critics using Eq.~\eqref{eq:td_target}
%         \State Update actor using Eq.~\eqref{eq:td3_bc}
%     \EndFor
% \EndFor
% \end{algorithmic}
% \end{algorithm}
%===============================================================================

\vspace{-1em}
\section{Experiments}
\begin{figure}[htbp]
    \centering
    \includegraphics[width=1\linewidth]{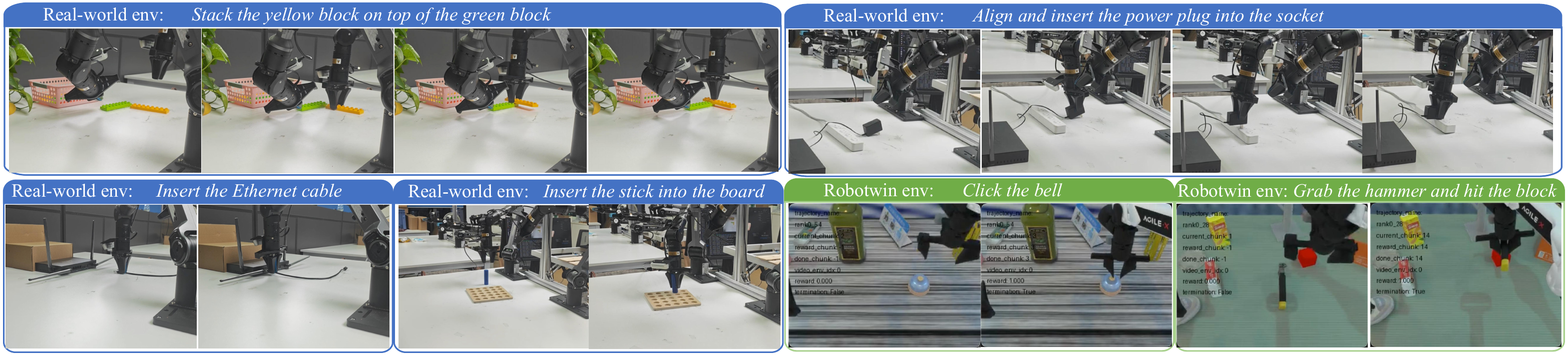}
    \caption{Representative rollouts of the evaluated tasks. Blue panels show four real-world precision manipulation tasks: Block Assembly, Power Plug Insertion, Ethernet Insertion, and Stick Insertion. Green panels show two RoboTwin simulation tasks: Click Bell and Beat Block Hammer. The task prompts are shown at the top of each panel.}
    \label{fig:experiment_show}
\end{figure}

\vspace{-1em}
We evaluate HALO-WA from three aspects, with representative real-world and simulation task rollouts shown in Fig.~\ref{fig:experiment_show}. First, we compare HALO-WA with representative baselines on four real-world precision manipulation tasks to verify whether it can improve the end-stage precision of WA models in real-world deployment. Second, we conduct key ablation studies on the Stick Insertion task to analyze the necessity of hybrid attention and WA latent-guided representation. Third, we perform two reproducible simulation experiments in RoboTwin, which also allow us to examine whether HALO-WA can improve online without human intervention. Due to space limitations, we report the core results in the main paper. More detailed real-world and simulation settings, as well as additional analyses, are provided in Appendix~A--G.
\vspace{-0.5em}
\subsection{Experimental Setup}
\vspace{-1em}

\vspace{-1em}
The real-world experiments are conducted on a Songling ALOHA dual-arm robot platform. The robot uses three RGB cameras as visual input, including one global-view camera and two wrist-view cameras. We adopt a cloud-robot collaborative online training setup. The cloud side uses two A800 GPUs: one for actor-critic updates and the other for rollout and policy execution. Before online training, we warm up the actor using supervised fine-tuning (SFT) with a BC loss so that it can roughly follow the WA reference actions, preventing large random action deviations at the early stage of online learning. During execution, the operator can intervene in real time, and the system stores the actually executed actions into the replay buffer for subsequent online updates.

% We evaluate four real-world precision manipulation tasks: Stick Insertion, Ethernet Insertion, Block Assembly, and Power Plug Insertion. All tasks require accurate end-stage alignment, contact, or insertion. WA-base can usually complete coarse approaching and action planning, but often fails in the last few millimeters of precise alignment and contact. The online training time for the four tasks is 45, 60, 75, and 60 minutes, respectively.

We evaluate four real-world precision manipulation tasks: Stick Insertion, Ethernet Insertion, Block Assembly, and Power Plug Insertion. All tasks require accurate end-stage alignment, contact, or insertion. WA-base can usually complete coarse approaching and action planning, but often fails in the last few millimeters of precise alignment and contact. For each task, we conduct 35 evaluation trials to measure the final policy performance. The online training time for the four tasks is 45, 60, 75, and 60 minutes, respectively.

We compare HALO-WA with four baselines. WA-base directly executes the frozen WA model, and we use GigaWorld Policy as the base WA model. Probe-Learn-Distill learns action residual corrections on top of a frozen base policy. HG-DAgger performs interactive imitation learning using human intervention data. RL-token-like adapter compresses WA latent features into a compact token and trains a lightweight actor-critic policy. The evaluation metrics include success rate (SR), episode length (EL), and the intervention-rate curves during training.

\begin{table*}[t]
\centering
\small
\setlength{\tabcolsep}{4.5pt}
\renewcommand{\arraystretch}{1.12}
\caption{
Main real-world results on four precision manipulation tasks.}
\vspace{-1em}
\label{tab:main_results}
\resizebox{\textwidth}{!}{
\begin{tabular}{l ccccc ccccc}
\toprule
\multirow{2}{*}{Method}
& \multicolumn{5}{c}{Success Rate (\%)$\uparrow$}
& \multicolumn{5}{c}{Episode Length$\downarrow$} \\
\cmidrule(lr){2-6}
\cmidrule(lr){7-11}
& \makecell{Stick\\Insertion}
& \makecell{Ethernet\\Insertion}
& \makecell{Block\\Assembly}
& \makecell{Power Plug\\Insertion}
& Average
& \makecell{Stick\\Insertion}
& \makecell{Ethernet\\Insertion}
& \makecell{Block\\Assembly}
& \makecell{Power Plug\\Insertion}
& Average \\
\midrule

WA-base (SFT)
& 34.3 & 40.0 & 11.4 & 20.0 & 26.4
& 268.3 & 306.6 & 360.3 & 375.6 & 327.7 \\

Probe-Learn-Distill\cite{xiao2025self}
& 62.9 & 65.7 & 34.3 & 45.7 & 52.1
& 245.8 & 284.7 & 337.6 & 351.2 & 304.8 \\

HG-DAgger\cite{kelly2019hg}
& \underline{77.1} & 74.3 & 45.7 & 54.3 & 62.9
& \underline{228.9} & 272.4 & \underline{320.4} & 340.7 & 290.6 \\

RL-token-like\cite{xu2026rl}
& 74.3 & \underline{82.9} & \underline{51.4} & \underline{62.9} & \underline{67.9}
& 236.5 & \underline{260.8} & 330.6 & \underline{334.1} & \underline{290.5} \\

\midrule

\textbf{HALO-WA}
& \makecell{\textbf{94.3}\\[-1pt]{\scriptsize(+175.0\%)}}
& \makecell{\textbf{97.1}\\[-1pt]{\scriptsize(+142.9\%)}}
& \makecell{\textbf{71.4}\\[-1pt]{\scriptsize(+525.0\%)}}
& \makecell{\textbf{85.7}\\[-1pt]{\scriptsize(+328.6\%)}}
& \makecell{\textbf{87.1}\\[-1pt]{\scriptsize(+230.0\%)}}
& \makecell{\textbf{211.4}\\[-1pt]{\scriptsize(-21.2\%)}}
& \makecell{\textbf{243.2}\\[-1pt]{\scriptsize(-20.7\%)}}
& \makecell{\textbf{300.8}\\[-1pt]{\scriptsize(-16.5\%)}}
& \makecell{\textbf{312.5}\\[-1pt]{\scriptsize(-16.8\%)}}
& \makecell{\textbf{267.0}\\[-1pt]{\scriptsize(-18.5\%)}} \\

\bottomrule
\end{tabular}}
\end{table*}

\vspace{-1em}
\begin{figure}[htbp]
    \centering
    \includegraphics[width=1\linewidth]{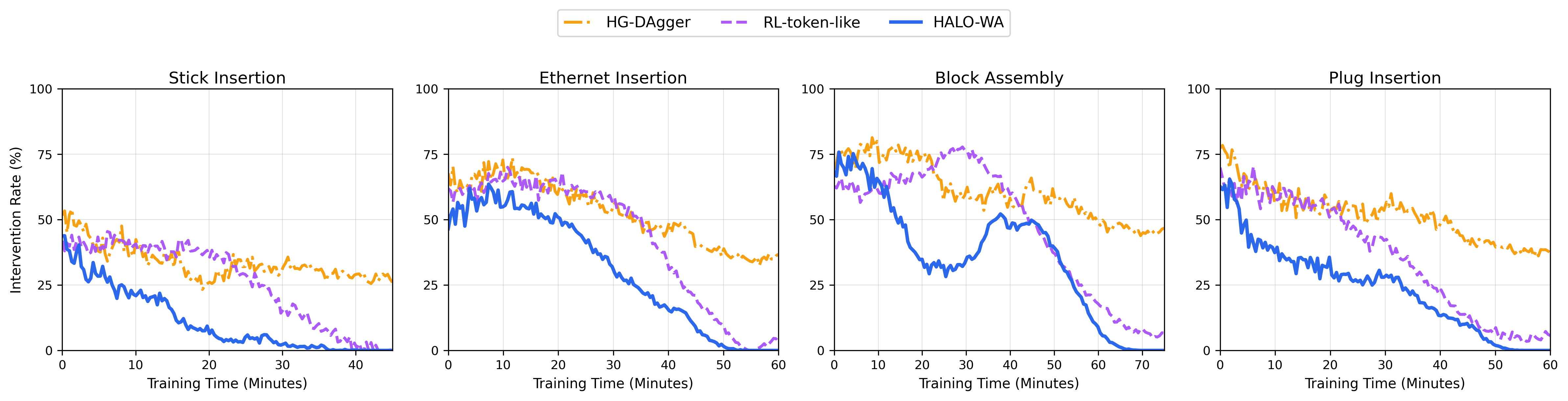}
    \caption{Intervention-rate curves during real-world online training. Lower intervention rates indicate reduced reliance on human correction. HALO-WA consistently decreases intervention across all four tasks and approaches zero in the later training stage.
}
    \label{fig:intervention}
    \vspace{-1.3em}
\end{figure}
% 

% \vspace{-0.5em}
\subsection{Main Real-World Results}
\vspace{-1em}
Table~\ref{tab:main_results} presents the main quantitative results on the four real-world tasks, and Fig.~\ref{fig:intervention} shows the intervention-rate curves during online training. HALO-WA achieves the best performance across all tasks, improving the average success rate from 26.4\% for WA-base to 87.1\% and reducing the average episode length from 327.7 to 267.0. This indicates that although the offline-trained WA model provides a useful coarse action prior, it still suffers from clear end-stage errors in real precision deployment. By learning online from real interaction, HALO-WA effectively bridges this precision deployment gap and turns the WA prior into reliable end-stage execution.
% \vspace{-2em}
Compared with other adaptation baselines, HALO-WA shows more consistent gains because it refines WA models at a more suitable level. Probe-Learn-Distill mainly performs local residual correction on top of the base action, which is limited for tasks requiring continuous alignment, contact feedback, and pose adjustment. HG-DAgger improves the policy with human intervention data, but pure imitation can be affected by inconsistent or suboptimal human takeover actions. Moreover, the comparison with HG-DAgger suggests that the policy improvement of HALO-WA does not rely solely on human intervention data. RL-token-like is the strongest baseline on average, reaching 67.9\% success rate, but compressing WA latents into a compact token may discard local spatial and action-generation details. In contrast, HALO-WA preserves distributed WA latent features and uses hybrid attention to read task-relevant information conditioned on the current state and WA reference action. This explains its stronger performance, especially on more complex tasks such as Block Assembly and Power Plug Insertion. Fig.~\ref{fig:intervention} further shows that HALO-WA rapidly reduces the intervention rate across all four tasks and approaches zero in the later stage, indicating that the policy gradually learns autonomous end-stage correction.

\vspace{-0.5em}
\subsection{Ablation Studies}
\vspace{-1em}
\begin{wraptable}{r}{0.32\linewidth}
\vspace{-8pt}
\centering
\footnotesize
\setlength{\tabcolsep}{3.5pt}
\renewcommand{\arraystretch}{1.02}
\caption{Ablation results.}
\label{tab:realworld_ablation_stick}
\begin{tabular}{lcc}
\toprule
Variant & SR$\uparrow$ & EL$\downarrow$ \\
\midrule
\multicolumn{3}{l}{\textit{Hybrid-Attention}} \\
Direct concat. & 28.6 & 276.4 \\
CNN encoder & 74.3 & 229.7 \\
Cross attn. & 57.1 & 246.9 \\
Self attn. & 62.9 & 239.1 \\
\textbf{Hybrid attn.} & \textbf{94.3} & \textbf{211.4} \\
\midrule
\multicolumn{3}{l}{\textit{Latent-Guided}} \\
Raw image enc. & 25.7 & 282.6 \\
WA inter. feat. & 57.1 & 248.3 \\
No latent & 22.9 & 287.5 \\
\textbf{WA latent feat.} & \textbf{94.3} & \textbf{211.4} \\
\bottomrule
\end{tabular}
\vspace{-8pt}
\end{wraptable}
% \vspace{-1em}

% \vspace{-1em}
We conduct two key ablation studies on the Stick Insertion task, as summarized in Table~\ref{tab:realworld_ablation_stick}. The first study evaluates the effect of hybrid attention. Direct concatenation achieves only 28.6\% success rate, suggesting that naively mixing high-dimensional WA latents, actions, and states makes it difficult to extract stable control cues. CNN\cite{lecun1998gradient} encoder improves the success rate to 74.3\%, indicating that WA latents contain useful spatial structure, but static compression lacks conditional interaction with the current action and state. Hybrid attention achieves the best performance, showing that combining self-attention for state-action structure with cross-attention for latent memory reading is more effective under limited real-world interaction data.

The second study evaluates the necessity of WA latent-guided representation. Raw image encoder and no latent achieve only 25.7\% and 22.9\% success rates, respectively, both lower than WA-base, showing that learning visual features from scratch or removing latent information is insufficient within short real-robot training. WA intermediate features reach 57.1\%, confirming that internal WA representations are useful, but they are less effective than the action-generation latents used by HALO-WA. These results show that the improvement mainly comes from effectively leveraging WA latent features with the proposed hybrid-attention design, rather than simply adding network capacity.
\vspace{-0.5em}
\subsection{Simulation Experiments}
% \vspace{-1em}

% \vspace{-1em}
\begin{wraptable}{r}{0.44\linewidth}
\vspace{-8pt}
\centering
\footnotesize
\setlength{\tabcolsep}{3.2pt}
\renewcommand{\arraystretch}{1.02}
\caption{RoboTwin simulation results.}
\label{tab:simulation_results}
\begin{tabular}{llcc}
\toprule
Task & Method & SR$\uparrow$ & EL$\downarrow$ \\
\midrule
\multirow{2}{*}{\makecell[l]{Click\\Bell}}
& WA-base & 54.0 & 151.4 \\
& \textbf{HALO-WA} & \textbf{97.0} & \textbf{100.3} \\
& & {\scriptsize(+79.6\%)} & {\scriptsize(-33.8\%)} \\
\midrule
\multirow{2}{*}{\makecell[l]{Beat Block\\Hammer}}
& WA-base & 22.0 & 312.5 \\
& \textbf{HALO-WA} & \textbf{45.0} & \textbf{213.7} \\
& & {\scriptsize(+104.5\%)} & {\scriptsize(-31.6\%)} \\
\bottomrule
\end{tabular}
\vspace{-8pt}
\end{wraptable}
\vspace{-1em}
To facilitate reproducibility, we further conduct two simulation experiments in RoboTwin that require precision manipulation: Click Bell and Beat Block Hammer. Unlike the real-world setting, the simulation setting does not provide human intervention as a strong corrective signal. This leads to longer trial-and-error learning. In addition, we do not warm up the actor in simulation, because without human intervention to break the WA behavior distribution, actor warm-up may overly constrain the policy to imitate the original WA actions and reduce exploration. For each simulation task, we evaluate each method over 100 trials and report the success rate as the percentage of successful trials.

Table~\ref{tab:simulation_results} presents the simulation results. On the Click Bell task, HALO-WA uses a single A800 GPU as the training device and improves the success rate from 54.0\% for WA-base to 97.0\% without human intervention. It also reduces the episode length from 151.4 to 100.3. On the longer-horizon and more complex Beat Block Hammer task, HALO-WA improves the success rate from 22.0\% to 45.0\%, corresponding to a relative improvement of 104.5\%, and reduces the episode length from 312.5 to 213.7. These results show that, even without human intervention, HALO-WA can still improve the performance of the frozen WA model through online adaptation in simulation.

% \begin{figure}[htbp]
%     \centering
%     \includegraphics[width=1\linewidth]{img/test4.png}
%     \caption{This is the caption of the figure.}
%     \label{fig:example}
% \end{figure}

% \begin{figure}[htbp]
%     \centering
%     \includegraphics[width=1\linewidth]{img/test5.png}
%     \caption{This is the caption of the figure.}
%     \label{fig:example}
% \end{figure}
\vspace{-1em}
\section{Limitations}
\label{sec:limitations}
\vspace{-1em}
Although HALO-WA significantly improves the deployment performance of WA models on real-world precision manipulation tasks, it still has several limitations. First, our real-world experiments mainly focus on insertion, alignment, and contact-rich assembly tasks. These tasks reflect the end-stage precision problem of WA models in real deployment, but they do not cover all robotic manipulation scenarios. Future work should further evaluate the generalization of HALO-WA across more object types, robot platforms, and long-horizon tasks. Second, HALO-WA relies on the pretrained WA model to provide reasonably effective action priors and latent features. When the reference action of the base WA model severely deviates from the task objective, or when its latent representation fails to encode sufficient task-relevant information, the correction capability of the lightweight adapter may be limited. Therefore, HALO-WA is more suitable for refining a WA policy that already has basic task competence, rather than solving tasks that the base model cannot perform at all. Finally, HALO-WA mainly uses visual latents, action priors, and online interaction feedback for precision correction, without explicitly modeling contact force, tactile feedback, or object dynamics. For more complex contact-rich tasks, incorporating force, tactile sensing, or finer-grained contact feedback may further improve robustness and safety in precision manipulation.
\vspace{-1em}
\section{Conclusion}
\label{sec:conclusion}
\vspace{-1em}
In this paper, we presented HALO-WA, a latent-guided online reinforcement learning framework for real-world deployment of world-action models. To address the end-stage alignment and insertion failures caused by calibration, perception, and contact-dynamics errors in real-world precision manipulation, HALO-WA leverages latent features and action priors from the WA generation process through a lightweight actor-critic adapter, enabling online action refinement without full-parameter online updates of the large WA backbone. We further designed a hybrid-attention structure that preserves the temporal consistency of action chunks while reading task-relevant information from WA latents conditioned on visual context and end-stage correction requirements. Real-world experiments show that HALO-WA substantially improves the deployment performance of WA-base on four precision manipulation tasks and achieves stable adaptation within a short online training time. Simulation experiments further demonstrate reproducible online improvement without human intervention. Overall, HALO-WA provides an efficient, lightweight, and online-adaptive approach for deploying large-scale world-action models in real-world precision manipulation.

%===============================================================================

\clearpage
% The acknowledgments are automatically included only in the final and preprint versions of the paper.
% \acknowledgments{If a paper is accepted, the final camera-ready version will (and probably should) include acknowledgments. All acknowledgments go at the end of the paper, including thanks to reviewers who gave useful comments, to colleagues who contributed to the ideas, and to funding agencies and corporate sponsors that provided financial support.}

%===============================================================================

\bibliography{example}  % .bib

% \maketitle

%===============================================================================

\section{Supplementary Material}

\subsection{Real-World Implementation Details}

The real-world experiments are conducted on an AgileX ALOHA dual-arm robot platform. The robot uses three RGB cameras as visual inputs, including one global-view camera and two wrist-view cameras. All images are resized to a unified resolution of $256 \times 256$ before being fed into the policy.

During deployment, the robot executes the action chunk predicted by either the frozen WA model or the HALO-WA adapter. When the operator observes an execution error or a potential failure, the operator can take over the robot in real time. During intervention, the system records the human-executed actions and stores them as online training data in the replay buffer.

To maintain a consistent chunk-level training format, after the intervention ends, the WA model re-infers a new reference action chunk from the current observation to complete the remaining steps of the action chunk. In this way, the replay buffer always stores samples with a standardized chunk-level structure.

HALO-WA adopts an asynchronous online training pipeline. The robot side is responsible for image acquisition, robot state reading, action execution, and intervention detection, while the training side performs WA inference, replay-buffer caching, and actor-critic updates. During online training, the WA backbone is always frozen, and only the lightweight HALO-WA adapter is updated.

\subsection{Baseline Details}

\paragraph{WA-base.}
WA-base uses the GigaWorld Policy trained through supervised fine-tuning as the base WA model. For each real-world task, we train a task-specific GigaWorld Policy separately. Specifically, we collect 100 real demonstration trajectories for each task and use the default training configuration of GigaWorld Policy for supervised fine-tuning. The checkpoint trained for 30k steps is used as the WA-base for the corresponding task. During the evaluation of WA-base, we directly execute the action chunks predicted by the frozen GigaWorld Policy, without enabling actor takeover. Therefore, this baseline is used to measure the original deployment performance of the task-level GigaWorld Policy trained through supervised fine-tuning on real-world precision manipulation tasks.

In addition, the 100 real demonstration trajectories collected for each task are not only used to train WA-base, but also used for offline actor initialization and online training. Specifically, we first use these trajectories to warm up the actor offline, so that the actor has a basic ability to follow the WA reference actions before online training starts, thereby avoiding unsafe or ineffective exploration caused by random actions in the early stage. Meanwhile, these trajectories are also added to the demonstration buffer, all treated as successful trajectories, and mixed with replay data generated from real online interactions at a 1:1 sampling ratio during online RL for actor-critic updates.

\paragraph{Probe-Learn-Distill.}
Probe-Learn-Distill learns a residual policy on top of the frozen base policy to locally correct the action output of the WA model. The improved behavior can be further distilled back into the policy model. This baseline represents the residual action-correction direction.

\paragraph{HG-DAgger.}
HG-DAgger updates the policy through imitation learning using human intervention data collected during online execution. This baseline evaluates whether supervised imitation from human corrective actions alone is sufficient to achieve performance comparable to HALO-WA.

\paragraph{RL-token-like.}
The RL-token-like follows the compact-token design of RLT. It compresses the WA latent features into a single compact representation and trains a lightweight actor-critic together with the WA reference action. This baseline is used to examine whether compressing distributed WA latent features into a compact token loses information required for precision manipulation.

\paragraph{HALO-WA.}
HALO-WA freezes the WA backbone and trains a lightweight actor-critic adapter using the WA reference action, WA latent features, and robot state. Before real-world online training, we use offline SFT to warm up the actor, so that the actor has a basic action-following capability.

\subsection{Real-World Task Details}

\paragraph{Stick Insertion.}
The robot is required to grasp a wooden stick and insert it into a hole on the target board. This task requires both position alignment and orientation correction at the final execution stage. In particular, the axis of the stick needs to be aligned with the direction of the hole. The online training time for this task is 45 minutes.
\begin{center}
\includegraphics[width=0.85\linewidth]{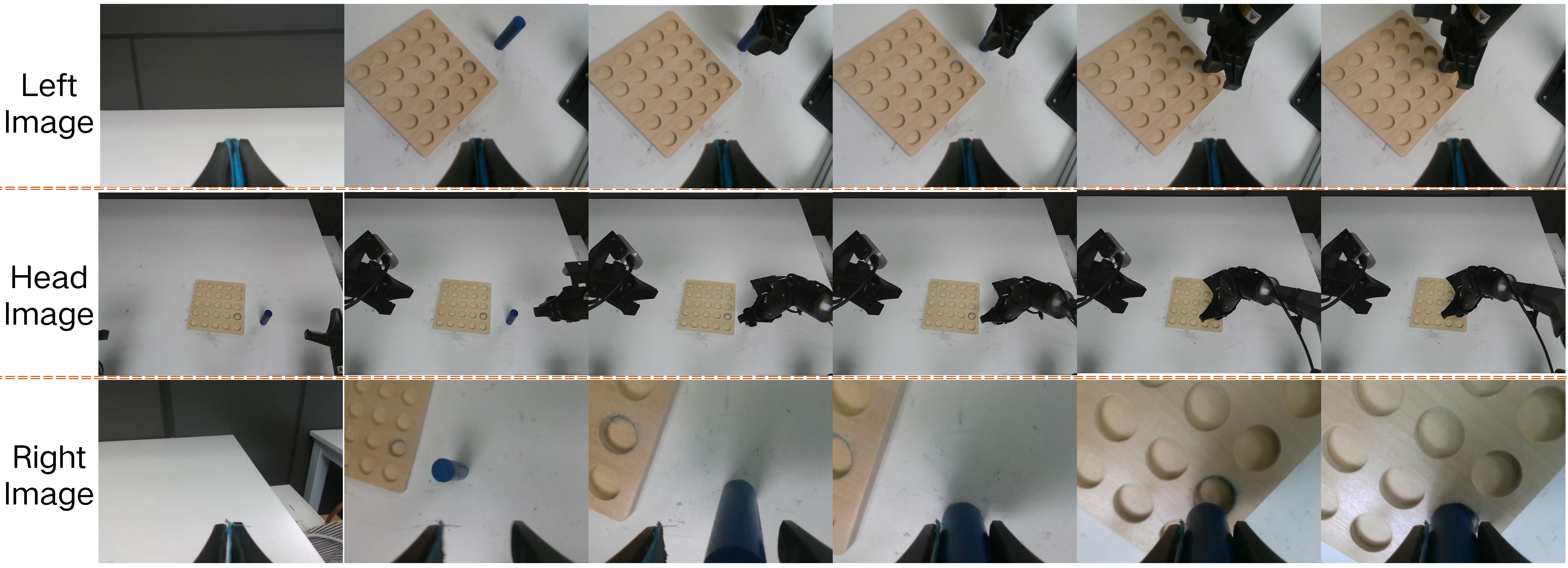}
\end{center}

\paragraph{Ethernet Insertion.}
The robot is required to insert an Ethernet connector into the corresponding port. This task requires accurate position, orientation, and insertion force control. The connector must be approximately aligned with the port direction and then pushed forward stably after contact. The online training time for this task is 60 minutes.
\begin{center}
\includegraphics[width=0.85\linewidth]{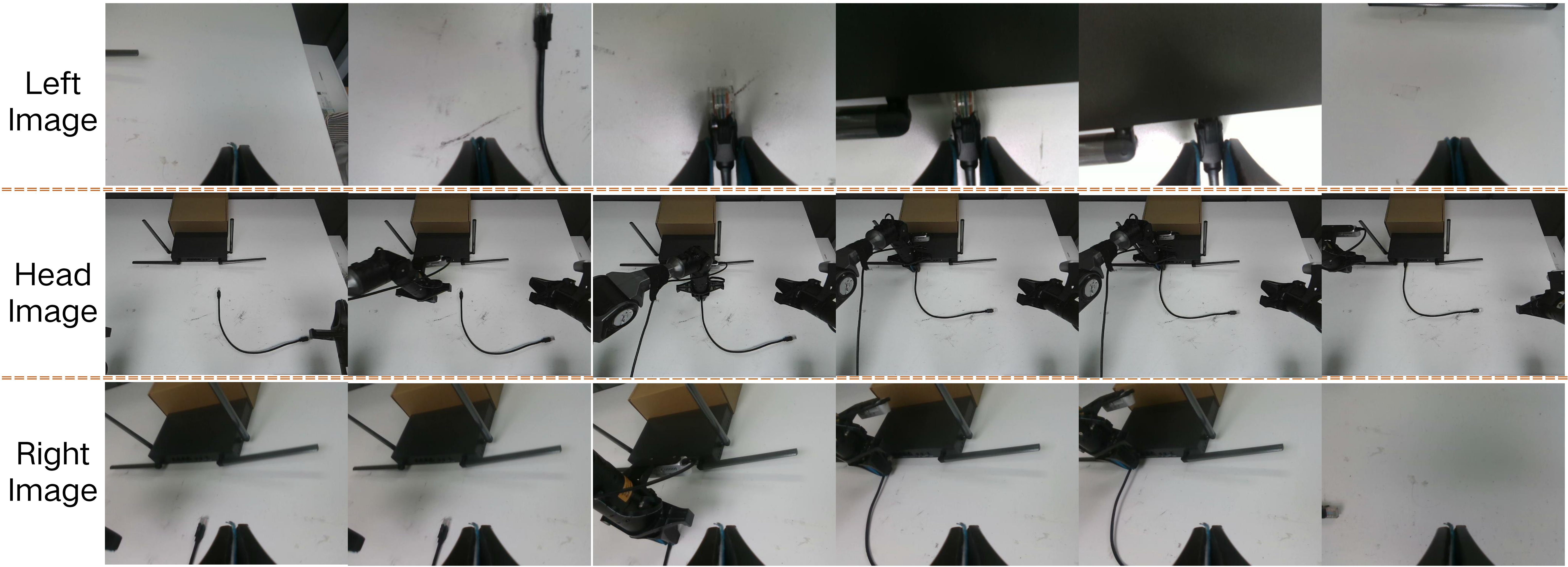}
\end{center}

\paragraph{Block Assembly.}
The robot is required to move the yellow block and assemble it with the green block at the specified position. Successful assembly requires the geometric interfaces of the two blocks to be aligned in both position and orientation. The online training time for this task is 75 minutes.
\begin{center}
\includegraphics[width=0.85\linewidth]{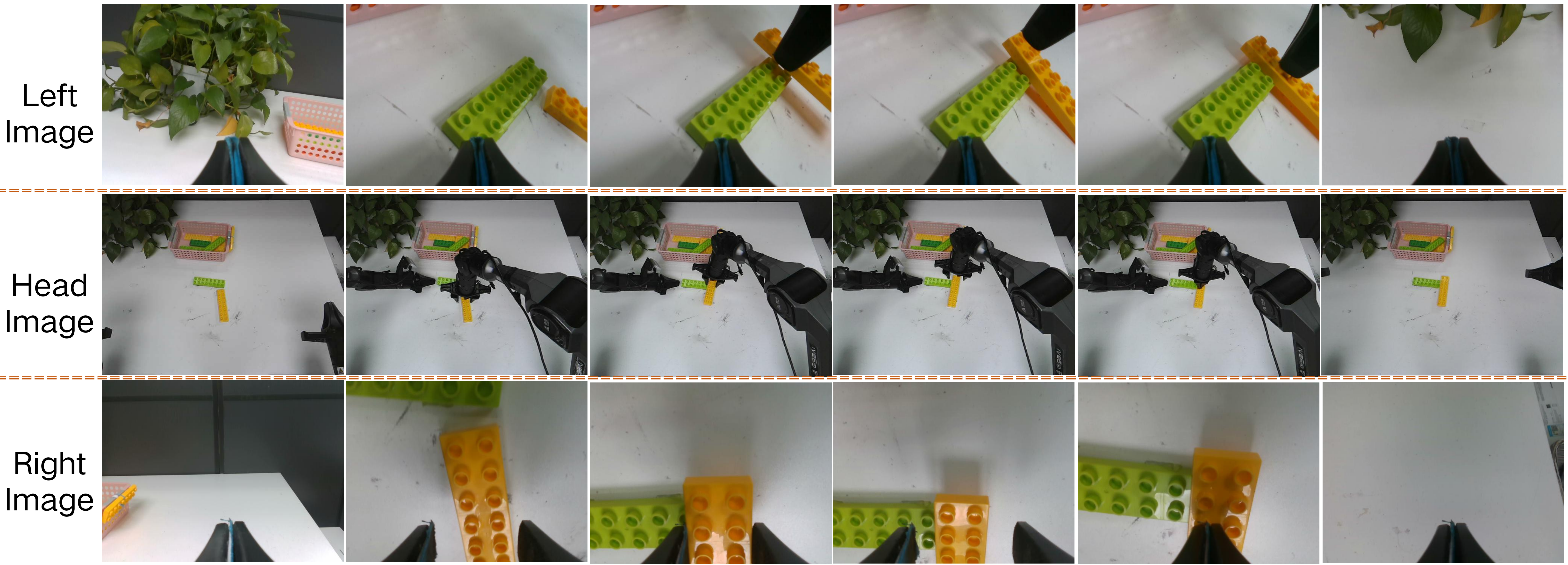}
\end{center}

\paragraph{Power Plug Insertion.}
The robot is required to grasp a power plug and insert it into a socket. The final insertion stage requires accurate plug orientation, socket localization, and stable contact behavior. The online training time for this task is 60 minutes.
\begin{center}
\includegraphics[width=0.85\linewidth]{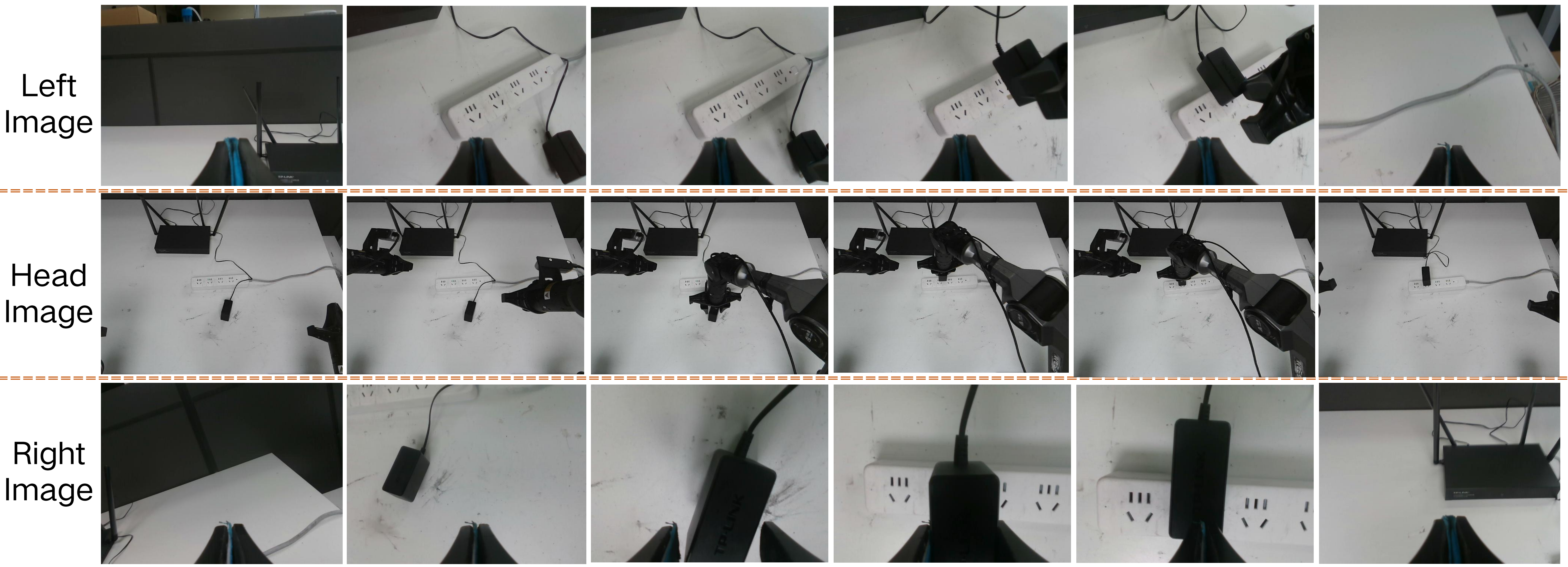}
\end{center}

\subsection{Additional Analysis of Real-World Results}

Compared with Probe-Learn-Distill, the advantage of HALO-WA mainly comes from a stronger action-refinement interface. Probe-Learn-Distill performs local residual correction on the output action of the base policy, which is closer to an action-level patch. However, for precision insertion and assembly tasks, success often depends on continuous end-effector alignment, contact adjustment, and orientation correction. Pure local residual correction is insufficient to capture this structured process. In contrast, HALO-WA directly takes the WA reference action, WA latent features, and robot state as inputs, and learns refined chunk-level actions through actor-critic training. Therefore, it can learn more continuous and structured precision-correction behaviors.

Compared with HG-DAgger, the advantage of HALO-WA is that it not only imitates human interventions but also further optimizes the policy using reinforcement-learning feedback. Human intervention data can improve the policy, but purely supervised imitation is sensitive to inconsistency and suboptimality in human corrective actions. In contact-rich tasks, the timing, angle, and magnitude of human interventions may vary substantially, and directly imitating these actions may introduce noise into the policy. HALO-WA treats human interventions as reliable behavioral data while using the critic to evaluate which corrections actually contribute to task completion. As a result, it can go beyond pure imitation while maintaining safe online adaptation.

Compared with the RL-token-like adapter, the advantage of HALO-WA directly validates the latent-guided design. The RL-token-like adapter compresses the WA latent features into a single compact token, which may discard local spatial information and action-generation details required for precision manipulation. In contrast, HALO-WA preserves the distributed WA latent features and uses hybrid attention to dynamically select task-relevant information conditioned on the current robot state and WA reference action. Therefore, HALO-WA achieves more reliable and efficient online adaptation.

\subsection{Ablation Variant Details}

\paragraph{Direct concatenation.}
This variant directly concatenates the latent features, WA reference action, and robot state, and feeds the concatenated representation into an MLP. It is used to evaluate whether simple feature concatenation is sufficient.

\paragraph{CNN encoder.}
This variant treats the WA latent as a spatial feature map and uses a CNN to compress it before feeding it into the actor-critic. It is used to evaluate whether local spatial feature extraction can replace the attention mechanism.

\paragraph{Cross attention.}
This variant uses the robot state and WA reference action as queries to retrieve relevant information from WA latent tokens. It is used to evaluate whether conditional latent reading improves precision correction.

\paragraph{Self attention.}
This variant tokenizes the latent features, reference action, and robot state, and then fuses them through self-attention. It is used to evaluate whether global token interaction is effective.

\paragraph{Hybrid attention.}
This is the full version of HALO-WA. It first models the relationship among state-action tokens and then uses cross-attention to select task-relevant information from the WA latent features. This design jointly models the action prior, the current robot state, and the WA latent representation.

\paragraph{Raw image encoder.}
This variant removes WA latent features and directly encodes raw images using a convolutional network before feeding the features into the actor-critic. It is used to evaluate whether ordinary visual features are sufficient for online precision adaptation.

\paragraph{WA intermediate features.}
This variant uses intermediate features from the WA model as the adapter input instead of the final VAE/world-action latent features. It is used to analyze how different internal representations of the WA model affect online control.

\paragraph{WA latent features.}
This variant uses the latent features from the WA generation process as the adapter input. It corresponds to the latent-guided design of HALO-WA and preserves the world-action representation that is directly related to action generation.

\paragraph{No latent.}
This variant removes the latent input and only uses the WA reference action and robot state. It is used to verify whether the final WA action alone contains sufficient information or whether additional latent guidance is necessary.

\subsection{Additional Ablation Discussion}

For the hybrid-attention ablation, direct concatenation performs the worst, indicating that directly stacking high-dimensional latent features, actions, and states makes it difficult for the actor-critic to extract useful information. The CNN encoder performs better, suggesting that the WA latent indeed contains useful spatial structure. However, CNN compression is mostly static and lacks conditional interaction with the WA reference action and robot state. Cross attention converges quickly, but it does not explicitly model the internal order of the action chunk or the state-action relationship. Self attention has strong representation capacity, but it introduces higher training cost and does not fully converge under the limited real-world training budget. Hybrid attention performs the best because it uses self-attention to preserve the action-chunk structure and cross-attention to conditionally retrieve task-relevant information from the WA latent features.

For the latent-guided ablation, both the raw image encoder and the no-latent variant perform worse than WA-base, indicating that it is unrealistic to relearn visual representations from raw images or to remove latent guidance entirely within a short real-world online training period. WA intermediate features outperform raw images and the no-latent variant, showing that internal WA representations do contain useful information. However, intermediate features are less directly tied to the action-generation process than WA latent features and therefore provide weaker support for final-stage contact and action refinement. The full HALO-WA method uses WA latent features, enabling the adapter to better judge whether the WA reference action is reliable and to learn accurate end-stage corrections.

\subsection{Simulation Implementation Details}

To facilitate reproducibility, we additionally evaluate HALO-WA in RoboTwin simulation. We select two tasks that require precision manipulation: Click Bell and Beat Block Hammer. Click Bell is a relatively simple end-effector contact task, where the robot needs to move accurately above the bell and press it. Beat Block Hammer is a longer-horizon and more complex task, where the robot must grasp the hammer handle at an appropriate position and accurately strike the target block. Object positions and initial states are randomized at the beginning of each episode, introducing distributional variations.

To construct WA-base, we train the base WA model with few-shot data, so that it has an initial ability to complete the task while still leaving clear room for online improvement. Unlike the real-world setting, RoboTwin does not provide human intervention as a strong corrective signal. Therefore, the actor relies on autonomous exploration. In the absence of human intervention, the actor takes over control throughout execution. Based on our empirical observations, we do not use WA-behavior warm-up for the actor in simulation. Instead, we randomly initialize the actor and directly start online RL. If the actor is warmed up by WA actions, it tends to stay close to the original WA policy distribution because no human intervention is available to break the WA-base behavior, making it more difficult to explore better actions.

The simulation experiments also follow an asynchronous online training pipeline. The simulated robot side runs the RoboTwin environment, while the training side performs actor-critic updates and policy synchronization. This reproduces the online-learning workflow of local execution and remote training.

\subsection{Actor-Critic Adapter Architecture}

The actor uses a cross-attention architecture with a hidden dimension of 512. The robot state and each WA reference action are first projected from 14 to 512 dimensions. Together with learnable action queries, they form 13 query tokens, including one state token and twelve action tokens. After self-attention and cross-attention with VAE visual latent tokens, the twelve action tokens are decoded by a shared token-wise MLP with dimensions
\[
512 \rightarrow 1024 \rightarrow 512 \rightarrow 14,
\]
producing a $12 \times 14$ action chunk.

The critic uses 25 query tokens, including one state token, twelve WA reference-action tokens, and twelve candidate-action tokens. All tokens are projected to 512 dimensions and cross-attend to the VAE visual latent tokens. The fused tokens are mean-pooled and passed into two independent Q-value heads. Each Q head is an MLP with dimensions
\[
512 \rightarrow 2048 \rightarrow 1024 \rightarrow 512 \rightarrow 1.
\]

\subsection{Training Hyperparameters}

For all tasks, the policy outputs an action chunk with length 12, and each action has 14 dimensions. Therefore, the policy predicts a $12 \times 14$ action chunk at each inference step. We use asynchronous rollout-training, where environment rollout and network training are performed in parallel. We also use a demonstration buffer for both real-world and simulation tasks. During online training, demonstration data and online replay data are mixed at a 1:1 sampling ratio in each training batch, i.e., 50\% demonstration data and 50\% online replay data. This mixed sampling strategy stabilizes early online training and prevents the actor from drifting too far from the WA reference behavior.

All experiments use sparse success/failure rewards instead of dense shaped rewards. In real-world experiments, the sparse reward is determined by task completion judged during execution, while in simulation experiments it is obtained from the environment success signal. A positive reward is assigned only when the task is successfully completed; otherwise, the reward remains zero. This setting allows us to evaluate whether HALO-WA can improve the frozen WA model without relying on task-specific dense reward engineering.

UTD ratio denotes the update-to-data ratio, i.e., the number of gradient update steps performed for each newly collected batch of rollout data. Unless otherwise specified, the common training settings are: UTD ratio of 5, critic-to-actor update ratio of 2, BC regularization coefficient of 50, global batch size of 64, micro-batch size of 4, actor learning rate of $2.5 \times 10^{-5}$, and critic learning rate of $3.0 \times 10^{-5}$.

The training hyperparameters for the four real-world tasks are summarized in Table~\ref{tab:real_world_training_hyperparameters}. The main task-dependent differences are the maximum episode length and the discount factor $\gamma$, which are adjusted according to the task horizon and manipulation complexity. Stick Insertion uses a shorter horizon of 192 steps with $\gamma=0.92$, Power Plug Insertion and Ethernet Insertion use a longer horizon of 384 steps with $\gamma=0.95$, and Block Assembly uses the longest horizon of 480 steps with $\gamma=0.98$.

\begin{table}[t]
\centering
\caption{Training hyperparameters for real-world tasks.}
\label{tab:real_world_training_hyperparameters}
\resizebox{\linewidth}{!}{
\begin{tabular}{lccccccccc}
\toprule
Task & Episode Length & $\gamma$ & Demo Ratio & Chunk Size & UTD & Critic/Actor & BC Coef. & Batch & Actor/Critic LR \\
\midrule
Stick Insertion & 192 & 0.92 & 0.5 & $12 \times 14$ & 5 & 2 & 50 & 64 & $2.5{\times}10^{-5}$ / $3.0{\times}10^{-5}$ \\
Power Plug Insertion & 384 & 0.95 & 0.5 & $12 \times 14$ & 5 & 2 & 50 & 64 & $2.5{\times}10^{-5}$ / $3.0{\times}10^{-5}$ \\
Block Assembly & 480 & 0.98 & 0.5 & $12 \times 14$ & 5 & 2 & 50 & 64 & $2.5{\times}10^{-5}$ / $3.0{\times}10^{-5}$ \\
Ethernet Insertion & 384 & 0.95 & 0.5 & $12 \times 14$ & 5 & 2 & 50 & 64 & $2.5{\times}10^{-5}$ / $3.0{\times}10^{-5}$ \\
\bottomrule
\end{tabular}}
\end{table}

The training hyperparameters for the RoboTwin simulation tasks are summarized in Table~\ref{tab:simulation_training_hyperparameters}. Both simulation tasks use the Aloha-AgileX embodiment, a maximum episode length of 192 steps, asynchronous rollout-training, sparse success/failure rewards, and a demonstration buffer with a demo sampling ratio of 0.5. The main difference between the two simulation tasks is the discount factor: Click Bell uses $\gamma=0.82$, while Beat Block Hammer uses $\gamma=0.92$.

\begin{table}[t]
\centering
\caption{Training hyperparameters for RoboTwin simulation tasks.}
\label{tab:simulation_training_hyperparameters}
\resizebox{\linewidth}{!}{
\begin{tabular}{lccccccccc}
\toprule
Task & Embodiment & Episode Length & $\gamma$ & Demo Ratio & Chunk Size & UTD & Critic/Actor & BC Coef. & Actor/Critic LR \\
\midrule
Click Bell & Aloha-AgileX & 192 & 0.82 & 0.5 & $12 \times 14$ & 5 & 2 & 50 & $2.5{\times}10^{-5}$ / $3.0{\times}10^{-5}$ \\
Beat Block Hammer & Aloha-AgileX & 192 & 0.92 & 0.5 & $12 \times 14$ & 5 & 2 & 50 & $2.5{\times}10^{-5}$ / $3.0{\times}10^{-5}$ \\
\bottomrule
\end{tabular}}
\end{table}

%===============================================================================

\end{document}